\documentclass{article}
\usepackage{ijcai22}

\usepackage{times}

\usepackage{soul}
\usepackage{url}
\usepackage[hidelinks]{hyperref}
\usepackage[utf8]{inputenc}
\usepackage[small]{caption}
\usepackage{graphicx}
\usepackage{amsmath}
\usepackage{booktabs}
\usepackage{subfig}
\usepackage{array}
\usepackage{mathtools}
\usepackage{booktabs}
\usepackage{amsfonts}
\usepackage{amsmath}
\usepackage{xspace}

\usepackage[switch]{lineno}
\let\oldequation\align
\let\oldendequation\endalign

\renewenvironment{align}
  {\linenomathNonumbers\oldequation}
  {\oldendequation\endlinenomath}

\makeatletter
\DeclareRobustCommand\onedot{\futurelet\@let@token\@onedot}
\def\@onedot{\ifx\@let@token.\else.\null\fi\xspace}

\def\eg{\emph{e.g}\onedot} 
\def\ie{\emph{i.e}\onedot} 
 
\def\etc{\emph{etc}\onedot}

\makeatother

\DeclareMathOperator*{\argmin}{arg\,min}
\usepackage{dsfont}

\newcommand\nnfootnote[1]{%
  \begin{NoHyper}
  \renewcommand\thefootnote{}\footnote{#1}%
  \addtocounter{footnote}{-1}%
  \end{NoHyper}
}

\title{Multi-level Consistency Learning for Semi-supervised Domain Adaptation}

\author{
Zizheng Yan$^{1*}$
\and
Yushuang Wu$^{1*}$
\and
Guanbin Li$^{2,1\dag}$
\and
Yipeng Qin$^{3}$ \\
Xiaoguang Han$^{1}$
\and
Shuguang Cui$^{1}$
\affiliations
$^1$Shenzhen Research Institute of Big Data, The Chinese University of Hong Kong, Shenzhen \\
$^2$Sun Yat-sen University \\
$^3$Cardiff University\\
\emails
\{zizhengyan@link, yushuangwu@link, hanxiaoguang@, shuguangcui@\}.cuhk.edu.cn,
liguanbin@mail.sysu.edu.cn, qiny16@cardiff.ac.uk
}

\begin{document}

\maketitle

\begin{abstract}
Semi-supervised domain adaptation (SSDA) aims to apply knowledge learned from a fully labeled source domain to a scarcely labeled target domain. 
In this paper, we propose a Multi-level Consistency Learning (MCL) framework for SSDA. 
Specifically, our MCL regularizes the consistency of different views of target domain samples at three levels: (i) at inter-domain level, we robustly and accurately align the source and target domains using a prototype-based optimal transport method that utilizes the pros and cons of different views of target samples; (ii) at intra-domain level, we facilitate the learning of both discriminative and compact target feature representations by proposing a novel class-wise contrastive clustering loss; (iii) at sample level, we follow standard practice and improve the prediction accuracy by conducting a consistency-based self-training.
Empirically, we verified the effectiveness of our MCL framework on three popular SSDA benchmarks, \ie, VisDA2017, DomainNet, and Office-Home datasets, and the experimental results demonstrate that our MCL framework achieves the state-of-the-art performance. Code is available at \url{https://github.com/chester256/MCL}.
\nnfootnote{$^*$Equal contribution}
\nnfootnote{$^\dag$Corresponding author}
\end{abstract}

\section{Introduction}

\begin{figure}[t]
    \centering
    \includegraphics[width=8cm]{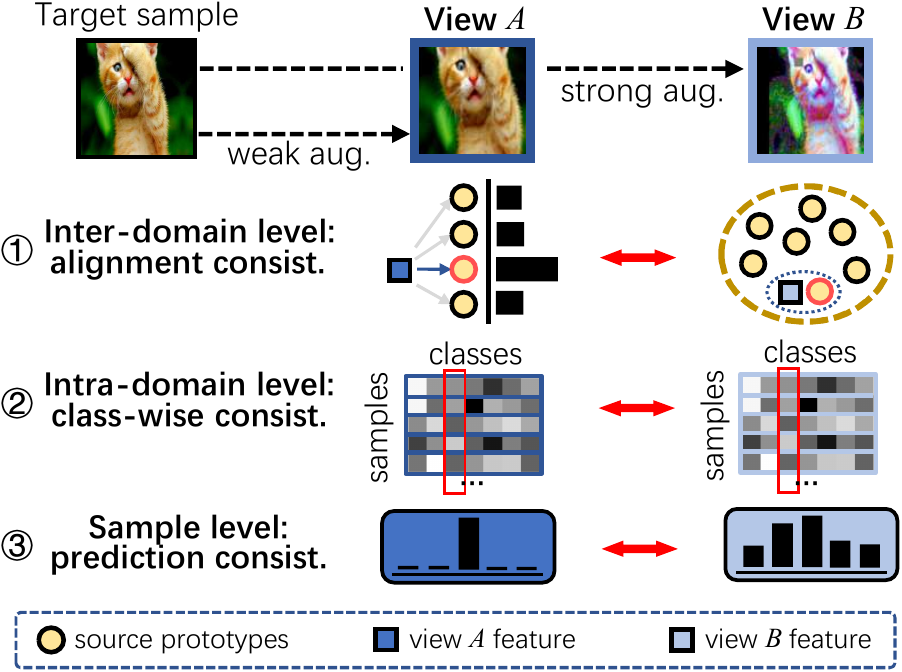}\\
    \caption{
    Overview of our Multi-level Consistency Learning (MCL) approach. Our MCL considers the consistency learning at three levels: (1) inter-domain level,  2) intra-domain level, and (3) sample level. We use various gray scales to indicate the confidence score of each sample to be classified as one class, and ``aug." and ``consist." represent augmentation and consistency, respectively. 
    }
    \label{teaser}
\end{figure}

Semi-supervised Domain Adaptation (SSDA) has attracted lots of attention due to its promising performance compared with Unsupervised Domain Adaptation (UDA).
Since only a few labeled target samples are available in SSDA, it can be viewed as a combination of UDA and Semi-supervised Learning (SSL) subtasks. 
Thus, it is straightforward to borrow ideas from SSL to address SSDA.
For example, \cite{yang2020mico} use mixup and co-training to boost the performance of SSDA. More recently, \cite{li2021cross} utilize consistency regularization~\cite{xie2019unsupervised,sohn2020fixmatch} and achieve state-of-the-art performance on several SSDA benchmarks. 

Although~\cite{li2021cross} show that consistency regularization benefits SSDA, it has only been naively applied at the sample level, \ie, enforcing the prediction of an unlabeled target sample to be consistent with those of its different views (obtained via augmentations). 
However, in SSDA, it is also important to attend the inter-domain knowledge transfer to better leverage the source supervision and intra-domain representation learning for both compact and discriminative class clusters.
Therefore, to fully utilize consistency regularization in SSDA, we propose a Multi-level Consistency Learning (MCL) framework, as shown in Fig.~\ref{teaser}, which regularizes the model with (i) inter-domain level alignment consistency, (ii) intra-domain level class-wise assignment consistency, and (iii) sample level prediction consistency. 

At the inter-domain level, we use consistency regularization to robustly and accurately align the source and target domain. We regularize the consistency between one view~(weakly augmented target samples) to source mapping and another view~(strongly augmented target samples) to source mapping.
Specifically, we first solve the optimal transport mapping between one view of the target samples and the source prototypes, and leverage the solved mapping to cluster another view to the source prototypes.
At the intra-domain level, we propose a novel class-wise contrastive clustering loss function to help the network learn both discriminative and compact feature representations in the target domain. This is implemented by computing the class-wise cross-correlation between the predictions from two views (target samples) in a mini-batch, and increasing the correlation of two views in the same class while reducing those from different classes. 
At the sample level, we follow standard practice and enforce the predictions of the weakly-augmented and the strongly-augmented views to be consistent.
In this way, the sample-level information can be fully probed to learn better feature representations for each single unlabeled target sample. 
With our multi-level consistency learning framework, the model is encouraged to capture complementary and comprehensive information at different levels, thereby learning more well-aligned and discriminative target features.

In summary, our work has the following contributions: (1) We propose a novel Multi-level Consistency Learning (MCL) framework for SSDA tasks. Our framework benefits SSDA by fully utilizing the potential of consistency regularization at three levels: the inter-domain level, the intra-domain level, and the sample level. (2) Our inter-domain level consistency learning method consists of a novel mapping and clustering strategy, and a novel prototype-based optimal transport method, which yields more robust and accurate domain alignment. (3) Our intra-domain level consistency learning method consists of a novel class-wise contrastive clustering loss, which helps the network learn more compact intra-class and discriminative inter-class target features. (4) Extensive experiments on SSDA benchmarks demonstrate that our MCL framework achieves a new state-of-the-art result.

\section{Related Work}

\paragraph{Semi-supervised Domain Adaptation (SSDA)}
Unlike UDA, SSDA assumes that there exist several labeled samples in the target domain, \eg 3 shots per class.
Many works have recently been proposed to tackle the SSDA problem~\cite{saito2019semi,yang2020mico,qin2021contradictory,li2021semi,kim2020attract,li2021cross,jiang2020bidirectional,yao2022enhancing,singh2021clda}. 
\cite{saito2019semi} first proposes to solve the SSDA problem by min-maximizing the entropy of target unlabeled data.
\cite{jiang2020bidirectional} train the network with virtual adversarial perturbations so that the network can generate robust features. 
\cite{yang2020mico} decompose SSDA into an inter-domain UDA subtask and an intra-domain SSL subtask, and utilize co-training to make the two networks teach each other and learn complementary information.
\cite{qin2021contradictory} use two classifiers to learn scattered source features and compact target features, thereby enclosing target features by the expanded boundary of source features.
\cite{li2021cross} propose to reduce the intra-domain gap by gradually clustering similar target samples together while pushing dissimilar samples away. 
In this work, instead of focusing on either inter-domain level or intra-domain level, we address both of them simultaneously so that the learning of both levels will benefit each other to learn more representative target features. 

\begin{figure*}[ht!]
    \centering
    \includegraphics[width=17.6cm]{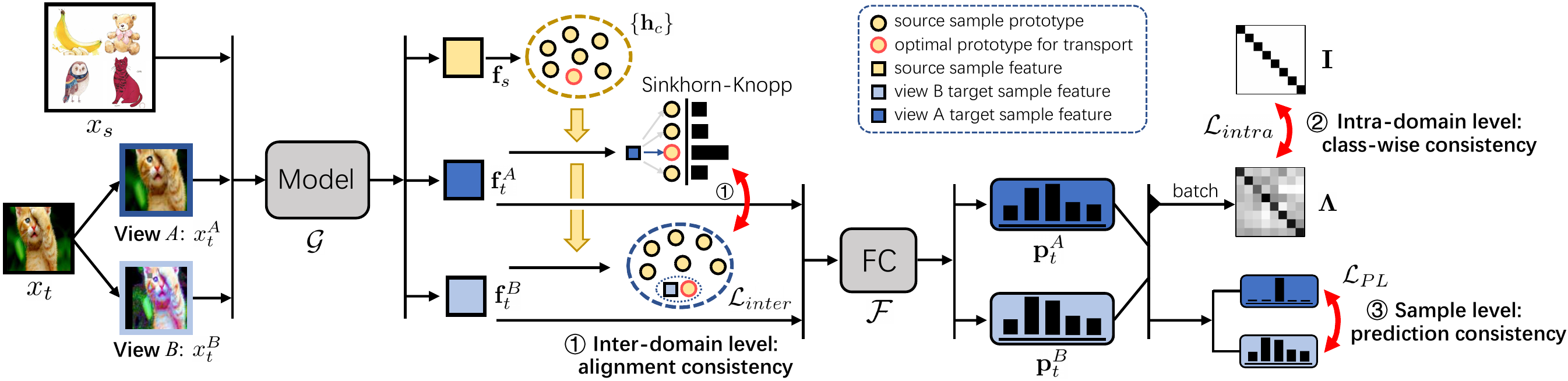}\\
    \caption{Illustration of the proposed MCL framework. Samples from the target domain are first augmented weakly and strongly to create two views $A$ and $B$. Consistency learning is then built upon these two views at three levels: (1) inter-domain level, where the alignment from two views' target features to source prototypes are enforced to be consistent; (2) intra-domain level, where class-wise consistency is kept in the target domain by enforcing the cross-correlation matrix of two views' batch-wise predictions to be close to an identity matrix; (3) sample level, where each sample has the consistent prediction for two views. }
    \label{overview}
\end{figure*}

\paragraph{Consistency Regularization}
Consistency regularization is a widely used technique in semi-supervised learning~\cite{sohn2020fixmatch} and self-supervised learning~\cite{chen2020simple}. 
In semi-supervised learning~\cite{sohn2020fixmatch}, enforcing the consistency among different augmentations of unlabeled samples can effectively boost the model performance. 
Recent works~\cite{chen2020simple} have shown that the consistency regularization objective can guide the model to learn meaningful representations in an unsupervised manner without any annotations, 
which inspires its application in label-scarce learning scenarios, \eg~domain adaptation, semi-supervised learning. Recently, some works have demonstrated that consistency regularization can also be utilized in domain adaptation tasks~\cite{french2017self}. 
Although our work is also built upon consistency regularization, rather than regularizing the sample-wise predictions only, we additionally regularize the consistency at both inter-domain and intra-domain levels.

\section{Methodology}
In this section, we first present the preliminaries including the definition of SSDA and associated notations, and then introduce the proposed Multi-level Consistency Learning (MCL) framework in detail.
An outline of our MCL framework is illustrated in Fig.~\ref{overview}.

\subsection{Preliminaries}
In the semi-supervised domain adaptation (SSDA) problem, the source domain $\mathcal{D}^s = \{ \mathcal{X}_s, \mathcal{Y}_s \}$ is fully labeled, and the target domain $\mathcal{D}^t = \{ \mathcal{X}_t, \mathcal{Y}_{tl}\}$ contains a few labels $\mathcal{Y}_{tl}$ for partial samples $\mathcal{X}_{tl}$, so $\mathcal{D}^t = \{ \mathcal{X}_{tl}, \mathcal{Y}_{tl}\} \cup  \{ \mathcal{X}_{tu}\}$, where $\{ \mathcal{X}_{tu}\}$ is the unlabeled target sample set. SSDA methods aim to learn a classifier with high accuracy on the target domain using the rich data-label pairs $\{ \mathcal{X}_s, \mathcal{Y}_s \}$ and few ones $\{ \mathcal{X}_{tl}, \mathcal{Y}_{tl}\}$. 
We assume that (i) the source and target domains share the same label space of $C$ classes; 
(ii) there are only a few labels for each class in the target domain, \eg~one label (1-shot) or three labels (3-shot).

SSDA model usually consists of two components, a feature extractor $\mathcal{G}$ and a classifier $\mathcal{F}$, parameterized by $\theta_\mathcal{G}$ and $\theta_\mathcal{F}$, respectively. 
Following \cite{sohn2020fixmatch}, we first generate two views for each target sample $x_{t} \in \mathcal{X}_t$, denoted as $x_t^A$ and $x_t^B$, through the weak (view $A$) and strong (view $B$) augmentation, respectively. 
The two views are then fed into $\mathcal{G}$ to obtain their feature representations $\mathbf{f}_t^A, \mathbf{f}_t^B \in \mathbb{R}^{d}$, which finally go through $\mathcal{F}$ to get the probabilistic prediction $\mathbf{p}_t^A, \mathbf{p}_t^B \in \mathbb{R}^{C}$. This process can be formulated as:
\begin{align}
    \mathbf{p}_t^v = \sigma \Big(\mathcal{F} \big( \mathbf{f}_t^v \big) \Big) = \sigma \Big( \mathcal{F} \big(\mathcal{G}(x_t^v) \big) \Big), ~v=A,B,
\end{align}
where $\sigma(\cdot)$ is the softmax function. 
Similarly, in the source branch, each source sample $x_s$ is fed into the same feature extractor $\mathcal{G}$ to generate $\mathbf{f}_s \in \mathbb{R}^{d}$, and then into the classifier $\mathcal{F}$ to obtain the prediction $\mathbf{p}_s \in \mathbb{R}^{C}$.

To train the model, each sample-label pair $(x,y)$ (from either the source or target domain) can be used to minimize a standard cross-entropy (CE) classification loss as follows:
\begin{align}
    \min_{\theta_\mathcal{G}, \theta_\mathcal{F}} \mathcal{L}_{CE} = -\sum_{(x,y)} y\log\big( p(x) \big),
    \label{eq:ce}
\end{align}
where $p(x)$ is the prediction of sample $x$ through $\mathcal{G}$ and $\mathcal{F}$. 
As detailed below, to fully utilize the unlabeled target data $\mathcal{X}_{tu}$, our MCL framework regularizes the consistencies in SSDA at three levels.

\subsection{Inter-domain Alignment Consistency}
At the inter-domain level, we propose a novel consistency regularization method to robustly and accurately align samples from the target domain to those of the source domain, so that the knowledge in the source domain can be better transferred into the target domain.
To obtain such a high-quality alignment, we compute the mapping plan from two views of the target sample to the source sample, and leverage the consistency between the two mappings to guide the learning process. Specifically, in each iteration, we first compute the mapping from the view $A$ of a target sample $x_t^A$ to a source sample $x_s$, according to which the representations of its view $B$, $x_t^B$ and $x_s$ are clustered closer. 
Accordingly, our method alternates the following two steps throughout the training:

\paragraph{Step 1. View-A to Source Mapping.}
We formulate the domain alignment as an optimal transport problem~\cite{courty2016OTDA}, which aims
to find the optimal sample-level coupling plan $\gamma^*$ between source samples and view-$A$ target samples 
by minimizing the total transport cost:
\begin{align}
    \gamma^* &= \argmin_{\gamma \in \mathbf{\Gamma}} \langle \gamma, ~\mathbf{C}^A\rangle_F, \label{eq:discrete_ot}\\
     \mathbf{\Gamma} &= \{ \gamma \in (\mathbb{R}^+)^{n_s \times n_t} ~\big|~\gamma\mathbf{1}_{n_t} = \mathbf{\mu}_s, ~\gamma^\top\mathbf{1}_{n_s} = \mathbf{\mu}_t\},
\end{align}
where $\gamma^*_{i,j}$ denotes the transport plan between the $i$th and $j$th sample in the source and target domain, respectively, $\langle\cdot, \cdot \rangle_F$ denotes the Frobenius inner product, $\mathbf{1}_{d}$ is a $d$-dimensional vector of 1's, and $\mathbf{\mu}_s \in \mathbb{R}^{n_s}, \mathbf{\mu}_t \in \mathbb{R}^{n_t}$ are the empirical distributions of source and target samples, respectively, which are usually set as uniform distributions. $\mathbf{C}^A \in \mathbb{R}^{n \times n}$ is the cost function matrix to indicate the transport cost between each source-target sample pair, with each element $\mathbf{C}^A_{i, j}$ in the matrix is inversely proportional to the cosine distance between the $i$-th source feature $\mathbf{f}_{si}$ and the $j$-th view-$A$ target feature $\mathbf{f}_{tj}^A$, computed as:
\begin{equation}
    \mathbf{C}^A_{i, j} = 1 - \mathbf{f}_{si}^\top\mathbf{f}_{tj}^A,
    \label{eq:new_cost_matrix}
\end{equation}
where more similar a feature vector pair is, a smaller transport cost can be obtained. With the cost matrix, $\mathbf{C}^A$ computed, the total transport cost can be minimized to obtain the optimal coupling plan $\gamma^*$, of which each element indicates a correspondence score for alignment.

\paragraph{Step 2. View-B to Source Clustering.}
With the guidance of optimal coupling $\gamma^*$, the feature representation of each strongly-augmented view-$B$ sample can be clustered toward the source prototype. In this way, the learned view-$B$ features can have a consistent mapping plan with the weakly-augmented view $A$.  
Hence the objective of the inter-domain alignment consistency learning is to minimize the loss function as below:
\begin{align}
    \min_{\theta_{\mathcal{G}}} \mathcal{L}_{inter} &= \langle \gamma^*, ~ \mathbf{C}^B\rangle_F, \label{eq:step_b}\\
    \text{with}~~~\gamma^* &\coloneqq \argmin_{\gamma \in \mathbf{\Gamma}} \langle \gamma, ~\mathbf{C}^A\rangle_F, \label{eq:step_a} \\
    \mathbf{C}^B_{i, j} &\coloneqq 1 - \mathbf{f}_{si}^\top\mathbf{f}_{tj}^B,
    \label{eq:align}
\end{align}
so that $\mathcal{G}$ can generate well-aligned view-$B$ features that are close to the corresponding source sample feature. 

\paragraph{Prototype-based OT.}
To further improve the accuracy and robustness, we propose to replace the source samples used in our OT method with source {\it prototypes}. Our source prototypes bring two benefits: (i) their features are more representative; (ii) they always cover all classes and avoid the misalignment error caused by missing classes in a mini-batch of source samples.
To this end, for each class $c \in \{1,2,\cdots,C\}$, we compute its prototype $\mathbf{h}_c$ by averaging the feature representations of all source samples in $c$ and substitute all $\mathbf{f}_{si}$ with $\mathbf{h}_{i}, i \in \{1,2,\cdots,C\}$ in Eq.~\ref{eq:discrete_ot}~and~\ref{eq:new_cost_matrix}. In Step~2 of our method, after the update of $\mathcal{G}$, each $\mathbf{h}_i$ in $\{\mathbf{h}_c\}$ is updated with the batch mean $\hat{\mathbf{h}}_i$ and a momentum $m$ after each iteration:
\begin{align}
    \mathbf{h}_i \leftarrow m \mathbf{h}_i + (1 - m) \hat{\mathbf{h}}_i.
    \label{eq:update}
\end{align}
Thus, we alternate between these two steps during training: (i) with fixed neural network parameters, solve the optimal coupling $\gamma^*$ for view $A$~(Eq.~\ref{eq:step_a}), and (ii) with the optimal coupling fixed, optimize the alignment loss for view $B$~(Eq.~\ref{eq:step_b}).

\subsection{Intra-domain Class-wise Consistency}
To facilitate representation learning in the target domain, we incorporate consistency regularization into the intra-domain level.
Specifically, we aim to learn both compact and discriminative class-wise clusters in the target domain, \ie, the clusters of the same class from two views should be grouped, while those of different classes should be pushed away from each other. Existing unsupervised contrastive learning methods focus on sample-level representation learning, which still hardly learn highly discriminative clusters for each class~\cite{khosla2020supervised}.

Instead of the sample-wise contrastive learning, we propose a novel class-wise contrastive clustering loss function, by modeling the cluster of each class as all samples' classification confidence of this class in a batch, named as {\it batch-wise class assignment} (illustrated as the column bounded by a red rectangle in Fig~.\ref{teaser}). The class-wise contrastive clustering has two purposes, (i) keeping the consistency of two views for the same batch-wise class assignment (positive pairs), and (ii) enlarging the distance between the batch-wise assignment of different classes (negative pairs). 

Formally, given a batch of target samples' predictions  $\mathbf{P} \in \mathbb{R}^{n \times C}$, where $n$ is the mini-batch size and $\mathbf{P}_{i,j}$ represents the confidence of classifying the $i$th sample into the $j$th class. With the cluster assumption that each feature of samples in the same class should form a high-density cluster, we regard each column of $\mathbf{P}$ as the batch-wise assignment of one class. Although the columns of $\mathbf{P}$ will dynamically change with the iteration of mini-batches during training, the class assignment in different views of the same image should be similar, while dissimilar to other images. Therefore, we compute the cross-correlation matrix $\mathbf{\Lambda} \in \mathbb{R}^{C\times C}$ between two views $A$ and $B$ to measure the class-wise similarity, formulated as:
\begin{align}
    \mathbf{\Lambda}\coloneqq {\mathbf{P}^A}^\top\mathbf{P}^B,
\end{align}
where $\mathbf{P}^A$ and $\mathbf{P}^B$ are the batch prediction from view $A$ and $B$, respectively. $\mathbf{\Lambda}$ is an asymmetry matrix with each element $\mathbf{\Lambda}_{i,j}$ measuring the similarity between $\mathbf{P}^A$'s $i$th column and $\mathbf{P}^B$'s $j$th column. The objective is to increase the correlation values between two views of the same class assignment (diagonal values) while reducing ones for different classes' views (off-diagonal values), which yields the loss function of intra-domain prediction consistency learning: 
\begin{align}
   \mathcal{L}_{intra} = \frac{1}{2C}(\big\Vert\phi(\mathbf{\Lambda}) - \mathbf{I}\big\Vert_1 + \big\Vert\phi(\mathbf{\Lambda}^\top) - \mathbf{I}\big\Vert_1),
    \label{eq:intra_1}\end{align}
where $\phi(\cdot)$ is a normalization function to keep the row summation as 1, \eg, similar to MCC ~\cite{jin2020minimum}, divide the elements with its row summation, $\mathbf{I} \in \mathbb{R}^{C\times C}$ is an identity matrix, and $\Vert \cdot \Vert_1$ denotes the summation of the absolute values of the matrix. Note that the loss function is the summation of two terms for respectively normalizing the rows and columns of the correlation matrix $\mathbf{\Lambda}$. Such two normalization operations can alleviate the influence of class imbalance in a batch, \ie, the values in the class assignment may be too large or small according to the number of samples that belong to this class. Intuitively, the normalized row of correlation values can be regarded as the scores of matching one class assignment with each column in the other view to be a positive pair. Thus the dual design helps bi-directional matching score computation, \ie., the $i$th row in $\phi(\mathbf{\Lambda})$ represents the  matching scores from $\mathbf{P}^A$'s $i$th column to $\mathbf{P}^B$'s each column, while the one in $\phi(\mathbf{\Lambda}^\top)$ represents matching in the inverse direction.

Note that minimizing Eq.~\ref{eq:intra_1} has the following properties: i) The rows of $\mathbf{P}^{A}$ and $\mathbf{P}^{B}$ are encouraged to be sharper, which implements entropy minimization~\cite{grandvalet2005entmin} in label-scarce scenarios.
    ii) For each sample, the prediction of view $A$ and $B$ are encouraged to become similar, which coincides with the consistency regularization~\cite{chen2020simple} that helps the network to learn better representations.
    iii) It avoids a trivial solution that predicts all samples as the same class: the diagonal values of the normalized cross-correlation matrix are enforced to be close to 1, which punishes the trivial solution by a large loss.

\begin{table*}[t]
    \centering
    \label{res_domainnet}
    \scalebox{0.82}{
        \begin{tabular}{l|cccccccccccccccc}
        \toprule
        & \multicolumn{2}{c}{R $\rightarrow$ C} 
        & \multicolumn{2}{c}{R $\rightarrow$ P} 
        & \multicolumn{2}{c}{P $\rightarrow$ C}
        & \multicolumn{2}{c}{C $\rightarrow$ S} 
        & \multicolumn{2}{c}{S $\rightarrow$ P}
        & \multicolumn{2}{c}{R $\rightarrow$ S} 
        & \multicolumn{2}{c}{P $\rightarrow$ R}
        & \multicolumn{2}{c}{\textbf{Mean}} \\  
        \textbf{Method}
        & \small 1-shot & \small 3-shot
        & \small 1-shot & \small 3-shot
        & \small 1-shot & \small 3-shot
        & \small 1-shot & \small 3-shot
        & \small 1-shot & \small 3-shot
        & \small 1-shot & \small 3-shot
        & \small 1-shot & \small 3-shot
        & \small \textbf{1-shot} & \small \textbf{3-shot}\\
        \midrule
        S+T      & 55.6 & 60.0   & 60.6 & 62.2  & 56.8 & 59.4   & 50.8 & 55.0              & 56.0 & 59.5   & 46.3 & 50.1   & 71.8 & 73.9   & 56.9 & 60.0 \\
        DANN   & 58.2& 59.8 &61.4& 62.8 &56.3& 59.6& 52.8 &55.4 &57.4& 59.9& 52.2 &54.9& 70.3 & 72.2 & 58.4 & 60.7   \\
        ENT      &  65.2 &  71.0  & 65.9  & 69.2  & 65.4  & 71.1 &  54.6  & 60.0  & 59.7  & 62.1 &  52.1  & 61.1 &   75.0 & 78.6  & 62.6  & 67.6 \\
        MME      &  70.0 &  72.2  & 67.7 &  69.7 &  69.0  & 71.7 &  56.3 &  61.8  & 64.8  & 66.8  & 61.0 &  61.9 &  76.1  & 78.5 &  66.4  & 68.9 \\
        UODA  & 72.7  & 75.4  & 70.3 &  71.5 &  69.8  & 73.2 &  60.5  & 64.1 &  66.4  & 69.4  & 62.7 &  64.2 &  77.3 &  80.8 &  68.5  & 71.2 \\
        MetaMME &  -  & 73.5 &  - &  70.3 &  -  & 72.8  & -  & 62.8 &  -  & 68.0  & - &  63.8 &  -  & 79.2 &  - &  70.1 \\
        BiAT  & 73.0 &  74.9 &  68.0 &  68.8 &  71.6 &  74.6 &  57.9 &  61.5 &  63.9 &  67.5  & 58.5 &  62.1 &  77.0 &  78.6 &  67.1 &  69.7 \\
        APE  & 70.4 &  76.6  & 70.8  & 72.1 &  72.9 &  76.7 &  56.7 &  63.1 &  64.5 &  66.1  & 63.0  & 67.8 &  76.6  & 79.4 &  67.6  & 71.7 \\
        ATDOC &  74.9 & 76.9 &  71.3 &  72.5  & 72.8 &  74.2 & 65.6  & 66.7 & 68.7  & 70.8 & 65.2 &  64.6 &  81.2  & 81.2 &  71.4 & 72.4 \\
        STar &  74.1 &  77.1 &  71.3 &  73.2 &  71.0 &  75.8 &  63.5 &  67.8 &  66.1  & 69.2  & 64.1  & 67.9 &  80.0 &  81.2 &  70.0 &  73.2 \\
        CDAC &  77.4  & 79.6 &  74.2 &  75.1 &  75.5 &  \textbf{79.3}  & \textbf{67.6}  & 69.9 &  71.0 &  73.4  & 69.2  & \textbf{72.5} &  80.4 &  81.9 &  73.6 &  76.0 \\
        DECOTA & \textbf{79.1} & \textbf{80.4} & \textbf{74.9} & 75.2 & \textbf{76.9} & 78.7 & 65.1 & 68.6 & 72.0 & 72.7 & 69.7 & 71.9 & 79.6 & 81.5 & 73.9 &  75.6 \\
        \midrule
        \textbf{MCL}   & 77.4 & 79.4
                        & 74.6 & \textbf{76.3}
                        & 75.5 & 78.8    
                        & 66.4& \textbf{70.9}
                        & \textbf{74.0}& \textbf{74.7}
                        & \textbf{70.7}&  72.3
                        & \textbf{82.0}& \textbf{83.3}
                        & \textbf{74.4}& \textbf{76.5}\\
        \bottomrule
        \end{tabular}
    }
    \caption{ Accuracy (\%) on \textit{DomainNet} under the settings of 1-shot and 3-shot using ResNet34 as backbone networks. }
    \label{table:domainnet}
\end{table*}

\begin{table*}[t]
    \centering
    \label{res_officehome}
    \setlength\tabcolsep{8.6pt}
    \scalebox{0.82}{
        \begin{tabular}{l|ccccccccccccc}
        \toprule
        \textbf{Method}
        & \small R $\rightarrow$ C
        & \small R $\rightarrow$ P
        & \small R $\rightarrow$ A
        & \small P $\rightarrow$ R
        & \small P $\rightarrow$ C
        & \small P $\rightarrow$ A
        & \small A $\rightarrow$ P
        & \small A $\rightarrow$ C
        & \small A $\rightarrow$ R
        & \small C $\rightarrow$ R
        & \small C $\rightarrow$ A
        & \small C $\rightarrow$ P
        & \textbf{Mean} \\  
        \midrule
        \multicolumn{14}{c}{\textbf{One-shot}}\\
        \midrule
        S+T      & 52.1 & 78.6 & 66.2 & 74.4 & 48.3 & 57.2 & 69.8 & 50.9 & 73.8 & 70.0 & 56.3 & 68.1 & 63.8 \\
        DANN   & 53.1 & 74.8 & 64.5 & 68.4 & 51.9 & 55.7 & 67.9 & 52.3 & 73.9 & 69.2 & 54.1 & 66.8 & 62.7 \\
        ENT      & 53.6 & 81.9 & 70.4 & 79.9 & 51.9 & 63.0 & 75.0& 52.9 & 76.7 & 73.2 & 63.2 & 73.6 & 67.9 \\
        MME      & 61.9 & 82.8 & 71.2 & 79.2 & 57.4 & 64.7 & 75.5 & 59.6 & 77.8 & 74.8 & 65.7 & 74.5 & 70.4 \\
        APE  & 60.7 & 81.6 & 72.5 & 78.6 & 58.3 & 63.6 & 76.1 & 53.9 & 75.2 & 72.3 & 63.6 & 69.8 & 68.9 \\
        CDAC & 61.9 & 83.1 & 72.7 & 80.0 & 59.3 & 64.6 & 75.9 & 61.2 & 78.5 & 75.3 & 64.5 & 75.1 & 71.0 \\
        DECOTA & 56.0 & 79.4 & 71.3 & 76.9 & 48.8 & 60.0 & 68.5 & 42.1 & 72.6 & 70.7 & 60.3 & 70.4 & 64.8 \\
        \midrule
        \textbf{MCL} & \textbf{67.0} & \textbf{85.5} & \textbf{73.8} & \textbf{81.3} & \textbf{61.1} & \textbf{68.0} & \textbf{79.5} & \textbf{64.4} & \textbf{81.2} & \textbf{78.4} & \textbf{68.5} & \textbf{79.3} & \textbf{74.0} \\

        \midrule[0.7pt]
        \multicolumn{14}{c}{\textbf{Three-shot}}\\
        \midrule
        S+T      &  55.7  & 80.8  & 67.8 &  73.1 &  53.8 &  63.5 &  73.1 &  54.0  & 74.2 &  68.3  & 57.6 &  72.3  &  66.2 \\
        DANN      & 57.3  & 75.5 &  65.2 &  69.2 &  51.8 &  56.6 &  68.3  & 54.7  & 73.8 &  67.1 &  55.1 &  67.5 &  63.5  \\
        ENT      & 62.6  & 85.7 &  70.2 &  79.9  & 60.5  & 63.9 &  79.5 &  61.3 &  79.1 &  76.4  & 64.7 &  79.1  & 71.9 \\
        MME      &  64.6 &  85.5 &  71.3 &  80.1  & 64.6 &  65.5 &  79.0 &  63.6 &  79.7 &  76.6 &  67.2 &  79.3 &  73.1 \\
        APE      & 66.4 &  86.2 &  73.4 &  82.0 &  65.2 &  66.1  & 81.1 &  63.9 &  80.2 &  76.8 &  66.6 &  79.9 &  74.0 \\
        CDAC  &  67.8  & 85.6  & 72.2 &  81.9  & 67.0  & 67.5 &  80.3  & 65.9  & 80.6  & 80.2 &  67.4 &  81.4 &  74.2 \\
        DECOTA & \textbf{70.4} &  87.7 & 74.0&  82.1&  68.0 & 69.9 & 81.8 & 64.0 & 80.5&  79.0 & 68.0 & 83.2 & 75.7 \\
        \midrule
        \textbf{MCL} & 70.1 & \textbf{88.1} & \textbf{75.3} & \textbf{83.0} & \textbf{68.0} & \textbf{69.9} & \textbf{83.9} & \textbf{67.5} & \textbf{82.4} & \textbf{81.6} & \textbf{71.4} & \textbf{84.3} & \textbf{77.1} \\
        \bottomrule
        \end{tabular}
    }
    \caption{ Accuracy (\%) on \textit{Office-Home} under the settings of 1-shot and 3-shot using ResNet34 as backbone networks. }
    \label{table:office_home}
\end{table*}

\subsection{Sample Prediction Consistency}
Similar to~\cite{li2021cross}, we regularize the consistency at the sample level via pseudo labeling: the prediction of view $A$ of a target sample is set as the pseudo label of view $B$ of the target sample.
Specifically, for each sample $x_{tu}$, we compute the cross-entropy loss as:
\begin{align}
    \mathcal{L}_{PL} = -\sum_{i=1}^{C}\mathds{1}(p_i^A \ge \tau) \log (p_i^B),
    \label{eq:pl}
\end{align}
where $p_i^A$ and $p_i^B$ are the $i$th elements in the prediction of $x_{tu}$ from view $A$ and view $B$, respectively, $\mathds{1}(\cdot)$ is the indication function, and $\tau$ is the threshold of confidence score.

\subsection{Overall Objective Function}
Therefore, the overall objective can be formulated as:
\begin{align}
    \min_{\theta_{\mathcal{G}}, \theta_{\mathcal{F}}} \mathcal{L}_{total} = \mathcal{L}_{CE} + \mathcal{L}_{PL} + \lambda_1 \mathcal{L}_{inter} + \lambda_2 \mathcal{L}_{intra},
\end{align}
where $\lambda_1$ and $\lambda_2$ are the hyper-paramters to balance $\mathcal{L}_{inter}$ and $\mathcal{L}_{intra}$, respectively. 
\begin{table}[t]
    \centering
    \setlength\tabcolsep{20pt}
    \scalebox{0.82}{
        \begin{tabular}{l|cc}
        \toprule
        \textbf{Method}
         &  1-shot
        &  3-shot \\
        \midrule
        S+T & 60.2 & 64.6 \\
        ENT & 63.6 & 72.7 \\
        MME  & 68.7 & 70.9 \\
        APE & 78.9 & 81.0 \\
        CDAC & 69.9 & 80.6 \\
        DECOTA & 64.9 & 80.7 \\
        \midrule 
        \textbf{MCL} & \textbf{86.3} & \textbf{87.3}   \\
        \bottomrule
        \end{tabular}
    }
    \caption{Mean Class-wise Accuracy(\%) on \textit{VisDA2017} using ResNet34 as backbone networks.}
    \label{res_visda}
    \vspace{-3mm}
\end{table}

\section{Experiments}
\subsection{Experiment Setups}
\paragraph{Datasets.}
We evaluate our proposed MCL on several popular benchmark datasets, including \textit{VisDA2017}~\cite{peng2017visda}, \textit{DomainNet}~\cite{domainnet}, and \textit{Office-Home}~\cite{officehome}.
\textit{DomainNet} was first introduced as the multi-source domain adaptation benchmark, and it consists of 6 domains with 345 categories. Following recent works~\cite{saito2019semi,li2021cross}, the \textit{Real}, \textit{Clipart}, \textit{Painting}, \textit{Sketch} domains with 126 categories are selected for evaluations. 
\textit{VisDA2017} is a popular synthetic-to-real domain adaptation benchmark, which consists of 150k synthetic and 55k real images from 12 categories. 
\textit{Office-Home} is a popular domain adaptation benchmark, which consists of 4 domains (\textit{Real}, \textit{Clipart}, \textit{Product}, \textit{Art}) and 65 categories.
Following most recent works~\cite{yang2020mico,li2021cross}, we conduct 1-shot and 3-shot experiments on all datasets.
Note that we follow most UDA works~\cite{long2013transfer} to report the Mean Class-wise Accuracy (MCA) as the evaluation metric for VisDA2017 and overall accuracy for DomainNet and Office-Home datasets.

\vspace{-1mm}
\paragraph{Implementation details.}
Similar to~\cite{saito2019semi,li2021cross}, we use a ResNet34 as the backbone for most experiments. The backbone is pre-trained on Imagenet~\cite{deng2009imagenet}. The batch size, optimizer, feature size, learning rate, \etc are set as same as in~\cite{saito2019semi}. Similar to~\cite{li2021cross}, the threshold $\tau$~(Eq.~\ref{eq:pl}) is set as 0.95, and we use the softmax temperature $T$ to control the sharpness of the prediction for the thresholding operation~(1 for \textit{Domainnet}, 1.25 for \textit{Office-Home} and \textit{VisDA}). The loss weight balancing hyperparameters $\lambda_1$ is set as 1, and $\lambda_2$ is set to 1 for \textit{DomainNet}, 0.2 for \textit{Office-Home} , and 0.1 for \textit{VisDA}. We use RandomFlip and RandomCrop as the augmentation methods for view $A$ and RandAugment~\cite{cubuk2020randaugment} for view $B$. For the OT solver, we introduce soft regularizations to the OT objective to make it a convex optimization problem that can be efficiently solved by Sinkhorn-Knopp iterations~\cite{fatras2021unbalanced}. Moreover, the momentum $m$ used to update source prototypes is set to 0.9. We use Pytorch and POT\footnote{https://pythonot.github.io} to implement our MCL framework. 

\subsection{Comparative Results}
We compare our MCL framework with i) two baseline methods S+T and ENT~\cite{grandvalet2005entmin}; and ii) several existing methods, including DANN~\cite{dann},  MME~\cite{saito2019semi}, UODA~\cite{qin2021contradictory}, BiAT~\cite{jiang2020bidirectional}, MetaMME~\cite{metamme}, APE~\cite{kim2020attract}, ATDOC~\cite{liang2021domain}, STar~\cite{singh2021improving}, CDAC~\cite{li2021cross}, and DECOTA~\cite{yang2020mico}.
Between the two baselines, ``S+T" uses only source samples and labeled target samples in the training; ``ENT" uses entropy minimization for unlabeled samples~\cite{grandvalet2005entmin}. Note that all methods are implemented on a ResNet34 backbone for a fair comparison.

\paragraph{Domainnet.}
Results on the \textit{DomainNet} dataset are shown in Tab.~\ref{table:domainnet}. To our knowledge, the proposed MCL achieves a new state-of-the-art (SOTA) performance, 74.4\% and 76.5\% mean accuracy over 7 adaptation scenarios, under the 1-shot and 3-shot setting, respectively. 

\paragraph{Office-Home.}
Results on the \textit{Office-Home} dataset further validate the effectiveness of MCL. As shown in Tab.~\ref{table:office_home}, MCL consistently outperforms other methods on both the 1-shot and 3-shot settings, and on all 12 adaptation scenarios. The mean accuracy of MCL reaches 74.0\% and 77.1\%, which outperform the SOTA performance by around 3\% and 1.4\%, under the 1-shot and 3-shot settings, respectively. 

\paragraph{VisDA.}
As shown in Tab.~\ref{res_visda}, MCL outperforms all existing SOTA methods in both 1-shot and 3-shot settings. Our MCL achieves 86.3\% MCA under the 1-shot setting, and 87.3\% for the 3-shot setting, significantly surpassing existing methods. 

\begin{table}[t]
    \centering
    \scalebox{0.85}{
        \begin{tabular}{lccc|cccc}
        \toprule
         \multicolumn{4}{c}{} 
        & \multicolumn{2}{c}{\small A $\rightarrow$ R}
        & \multicolumn{2}{c}{\small C $\rightarrow$ P} \\ 
        & \small$\mathcal{L}_{inter}$
        & \small$\mathcal{L}_{intra}$
        & \small$\mathcal{L}_{PL}$
        & \small1-shot
        & \small3-shot 
        & \small1-shot
        & \small3-shot \\ 
        \midrule
        ($a$) &  &  &  & 73.8 & 74.2 & 68.1 & 72.3 \\ 
        ($b$) & \checkmark &  &  & 75.8 & 77.7 & 72.5 & 78.6 \\ 
        ($c$) & & \checkmark & & 79.3 & 81.1 & 76.0 & 79.4 \\
        ($d$) & & &\checkmark & 75.5 & 77.2 & 72.8 & 79.1 \\
        ($e$) & \checkmark & \checkmark & & 79.7 & 81.1 & 78.2 & 81.6 \\
        ($f$) & \checkmark & & \checkmark & 75.9 & 78.2 & 73.9 & 82.0 \\
        ($g$) & & \checkmark & \checkmark & 79.3 & 81.5 & 78.4 & 82.9 \\
        ($h$) &\checkmark&\checkmark&\checkmark& 81.2 & 82.4 & 79.3 & 84.3 \\
        \bottomrule
        \end{tabular}
    }
    \caption{ Ablation studies of MCL's different components. We report the Accuracy (\%) on \textit{Office-Home} of A $\rightarrow$ R and C $\rightarrow$ P under the settings of 1-shot and 3-shot using a ResNet34 backbone. }
    \label{res_ablation}
\end{table}
\begin{table}[t]
    \centering
    \scalebox{0.85}{
        \begin{tabular}{cc|cc|cc}
        \toprule
        \multicolumn{2}{c}{\textbf{OT}}  
        & \multicolumn{2}{c}{\textbf{CC}}
        & \multicolumn{2}{c}{\textbf{Accuracy}}\\
        
        Standard
        & Proto.
        & Sample
        & Class
        & C $\rightarrow$ S
        & P $\rightarrow$ R\\
        \midrule
        \checkmark &   & \checkmark &  & 66.5   & 80.0 \\
        \checkmark &   &  & \checkmark  & 67.4   & 81.2 \\
          & \checkmark & \checkmark &    & 69.2  & 81.9 \\
          & \checkmark &   & \checkmark & \textbf{70.9} & \textbf{83.3} \\
        \bottomrule
        \end{tabular}
    }
     \caption{ Framework design analysis. Accuracy (\%) on two adaptation scenarios of \textit{DomainNet}, C $\rightarrow$ S and P $\rightarrow$ R, under the 3-shot setting. ``\textbf{CC}" means contrastive clustering, ``proto" means using prototype-based OT, ``sample" and ``class" mean sample-wise and class-wise, respectively. }
     \label{res_analysis}
    \vspace{-2mm}
\end{table}

\subsection{Analysis}
\paragraph{Ablation studies.}
We conduct ablation studies on the \textit{Office-Home} of A $\rightarrow$ R and C $\rightarrow$ P, and under both 1-shot and 3-shot settings, as shown in Tab.~\ref{res_ablation}. Row (b)(c)(d) show that each component can yeild significant improvement. Row (e)(f)(g) show that each combination can still improve the performance, which indicates the versatility of the proposed modules. The best performance is achieved with all components activated.

\paragraph{Prototype-based OT.}
In the proposed MCL, prototype-based OT is adopted to compute the mapping plan between target samples and source prototypes, which generates more representative source features for transport to learn robust inter-domain alignment. In comparison, we also implement sample-wise OT on the \textit{DomainNet} dataset under the 3-shot setting, with other hyper-parameters the same as MCL. Evaluation results on two adaptation scenarios are listed in Tab.~\ref{res_analysis}, sample-wise OT (``Standard") results in a drop of 3.5\% (C $\rightarrow$ S) and 2.1\% (P $\rightarrow$ R), which verify our claim.

\paragraph{Class-wise contrastive clustering.}
In the design of class-wise contrastive clustering loss, 
an alternative is to compute the {\it sample-wise} correlation matrix $\mathbf{\Lambda}' = \mathbf{P}^{A}{\mathbf{P}^B}^\top \in \mathbb{R}^{n\times n}$ instead, where $n$ is the batch size. We claim that such a sample-wise operation is inferior, and experimental results on the \textit{DomainNet} dataset under the 3-shot setting support our argument, as listed in Tab.~\ref{res_analysis}. 
One potential reason might be that the class-wise inner product focuses on clustering each class for higher compactness, whereas the sample-wise one pays more attention to the correlation between individual samples, which is less effective when labels are scarce and compact representations are difficult to generate.

\section{Conclusion}
In this paper, we presented a novel Multi-level Consistency Learning (MCL) framework for the SSDA problem.
MCL consists of three levels of consistency regularization, (i) inter-domain alignment consistency for learning robust inter-domain alignment via a prototype-based optimal transport method, (ii) intra-domain class-wise consistency for learning both compact and discriminative class clusters, and (iii) sample-level prediction consistency to better leverage instance-wise information. Our MCL approach well leverages information at multiple levels, with extensive experiments and ablation studies solidly proving its superiority.

\section*{Acknowledgments}
The work was supported in part by Chinese Key-Area Research and Development program of Guangdong Province (2020B0101350001), by the Basic Research Project No.~HZQB-KCZYZ-2021067 of Hetao Shenzhen-HK S\&T Cooperation Zone, by National Key R\&D Program of China with grant No.~2018YFB1800800, by Shenzhen Outstanding Talents Training Fund 202002, by Guangdong Research Projects No.~2017ZT07X152 and No.~2019CX01X104, by the National Natural Science Foundation of China under Grant No.61976250 and No.U1811463, by the Guangzhou Science and technology project under Grant No.202102020633 and by Shenzhen Research Institute of Big Data. Thanks to the ITSO in CUHKSZ for their High-Performance Computing Services.
\newpage
\bibliographystyle{named}
\bibliography{ijcai22}

\end{document}